\documentclass[11pt]{article}
\usepackage{coling2020}
\usepackage{times}
\usepackage{url}
\usepackage{latexsym}

\usepackage[T1]{fontenc}
\usepackage{type1cm}
\usepackage[american]{babel}
\usepackage{url}
\usepackage{amssymb} 
\usepackage{amsmath}
\usepackage{multirow} 
\usepackage{xspace}
\DeclareMathAlphabet{\mathcalbf}{OMS}{pzc}{b}{n}
\usepackage{microtype}

\usepackage{booktabs}
\usepackage{multirow}
\usepackage{graphicx}
\DeclareGraphicsExtensions{.pdf,.ai,.jpg,.png}
\setkeys{Gin}{pagebox=artbox}

\graphicspath{{./figures/}}

\newcommand{\bsfigure}[3][]{%
	\begin{figure}[t]
		\centering
		\includegraphics[#1]{#2}
		\caption{#3}\label{#2}%
 	 \end{figure}
}

\newcommand{\hwfigure}[3][t!]{%
	\begin{figure*}[#1]
		\centering
		\includegraphics[scale=1.0]{#2}
    		\caption{#3}\label{#2}%
  	\end{figure*}
}

\RequirePackage{color}
\RequirePackage{soul}
\setstcolor{blue}

\definecolor{highlight1}{rgb}{0.95,0.95,0.95}
\definecolor{tgray}{rgb}{0.5,0.5,0.5}

\begin{document}

%
%
\blfootnote{
    %
    %
    %
    %
    \hspace{-0.65cm}  
    This work is licensed under a Creative Commons 
    Attribution 4.0 International License.
    License details:
    \url{http://creativecommons.org/licenses/by/4.0/}.
}

\title{Semi-Supervised Cleansing of Web Argument Corpora}

\author{Jonas Dorsch \\
  Faculty of Media, Webis Group \\
  Bauhaus-Universit\"at Weimar \\
  Weimar, Germany \\
  {\tt jonas.dorsch@uni-weimar.de} \\\And
  Henning Wachsmuth \\
  Department of Computer Science \\
  Paderborn University \\
  Paderborn, Germany \\
  {\tt henningw@upb.de} \\}

\date{}

\maketitle

\begin{abstract}
Debate portals and similar web platforms constitute one of the main text sources in computational argumentation research and its applications. While the corpora built upon these sources are rich of argumentatively relevant content and structure, they also include text that is irrelevant, or even detrimental, to their purpose. In this paper, we present a precision-oriented approach to detecting such irrelevant text in a semi-supervised way. Given a few seed examples, the approach automatically learns basic lexical patterns of relevance and irrelevance and then incrementally bootstraps new patterns from sentences matching the patterns. In the existing args.me corpus with 400k argumentative texts, our approach detects almost 87k irrelevant sentences, at a precision of 0.97 according to manual evaluation. With low effort, the approach can be adapted to other web argument corpora, providing a generic way to improve corpus quality.
\end{abstract}

\section{Introduction}
\label{sec:introduction}

Computational argumentation research lays the ground for applications that support opinion formation, including argument search engines~\cite{wachsmuth:2017e}, collective deliberation~\cite{uszkoreit:2017}, and debating technologies~\cite{toledo:2019}. Such applications rely on large pools of up-to-date arguments, which can hardly be found anywere but on the web. One of the most used web argument sources are debate portals where people jointly collect arguments or debate each other on defined issues. Debate portals, and similar web platforms, are rich of argumentatively relevant content and structure, including arguments as well as facts, background information, and similar. This enables researchers to crawl large-scale argument corpora in a distantly-supervised manner \cite{alkhatib:2016a}. 

However, the texts found on debate portals also comprise debate-specific language and boilerplate text that is likely to be irrelevant, if not even detrimental, to the mentioned applications.  In the text in Figure~\ref{example-2}, for instance, the author defines the debated issue (sentence \#2), states a thesis (\#3--5), and presents two arguments (\#6--8, \#9--13) --- all of which can be considered argumentatively relevant. In contrast, sentences \#1, \#14, and \#15 add nothing of importance, merely making meta-comments and expressing gratitude. In other cases, irrelevant text includes salutations, insults, purely rhetorical moves, and spam. As detailed in Section~\ref{sec:relatedwork}, finding such text differs from finding non-argumentative text segments, since the latter may still be relevant as context for the argumentative segments, as in the case of sentence~\#2 in Figure~\ref{example-2}. Many existing approaches relying on debate portals do not clean the crawled arguments from irrelevant text. Until now, for example, the argument search engine {\em args.me} \cite{wachsmuth:2017e} has just returned the full shown text as one pro argument for the query ``gay marriage''. This at least harms user experience, and it might even corrupt the support of opinion formation in some cases.

\bsfigure{example-2}{Example text taken from a debate portal. Sentences \#1, \#14, and \#15 can be considered irrelevant to the arguments made by the author. Our approach learns basic lexical patterns to detect such sentences, here shown bold and underlined. Italicized phrases indicate patterns in sentences learned to be relevant.}

In this paper, we study how to find irrelevant text in web arguments such as those from debate portals automatically, in order to clean respective corpora on this basis. In particular, we develop a semi-supervised learning approach that aims to detect as many irrelevant sentences as possible with very high precision, i.e., hardly any relevant sentence should be classified as irrelevant (Section~\ref{sec:approach}). Given a seed set of sentences, the approach learns basic lexical $n$-gram patterns that frequently match text in either relevant or irrelevant sentences, and it keeps all patterns with some minimum precision (estimated on all matching sentences). Based on all matching sentences in a given corpus, it then bootstraps new patterns, revises previous ones, and incrementally repeats the process. The final set of irrelevance patterns is used to cleanse the corpus.\,\,\,

We analyze our approach on the args.me corpus \cite{ajjour:2019a}, consisting of 387,606 arguments from four debate portals, more than any other available corpus to our knowledge (Section \ref{sec:data}). Exploring different types of lexical patterns, we find that word $n$-grams ignoring stopwords serve best to distinguish relevant from irrelevant sentences. From the most frequent such $n$-grams, we manually select a set of seed sentences. Then, we run the bootstrapping process, analyze the patterns found by the \nopagebreak{approach over} its different iterations, and evaluate its precision both in an automatic way and in a manual annotation study with three human annotators on 600 sentences (Section~\ref{sec:experiments}). At a Fleiss' $\kappa$ agreement of 0.50, our approach detects irrelevant sentences with a precision of 0.97, in total 86,916 of them in 68,814 arguments from the args.me corpus. We provide a cleaned version of the corpus to the community.%
\footnote{Both the original and the cleaned args.me corpus are found at: \url{https://webis.de/data.html#args-me-corpus}}

Finally, we discuss how to adopt our approach to improve the quality of web argument corpora, beyond the one studied (Section~\ref{sec:conclusion}). Altogether, the contribution of this paper is three-fold:
\begin{itemize}
\setlength{\itemsep}{0pt}
\item
A semi-supervised approach to detect argumentatively irrelevant sentences in web arguments.
\item
Several common lexical patterns of relevance and irrelevance in web arguments.
\item
A cleaned version of the largest available argument corpus, with notably less irrelevant text.
\end{itemize}

\section{Related Work}
\label{sec:relatedwork}

Initially, research on tasks such as argument mining has largely been carried out on small, well-curated collections of texts, including Wikipedia articles \cite{aharoni:2014}, student essays \cite{stab:2014a}, pure arguments~\cite{peldszus:2015}, and presidential debates~\cite{lawrence:2017}. Major real-world applications of computational argumentation, however, need to scale up to web contexts to fulfill their purpose. This includes search engines that oppose pro and con arguments on controversial issues \cite{wachsmuth:2017e}, technologies that debate humans \cite{toledo:2019}, and more. 

To obtain web arguments, many works have relied on crawled debate portals and similar web platforms, often in a distant-supervision manner where argumentative structure and similar annotations are directly derived from available meta-information \cite{alkhatib:2016a}. Corpora have been built in such a way based on several debate portals, including {\em 4forums.com} \cite{walker:2012}, {\em idebate.org} \cite{cabrio:2012}, {\em createdebate.com} \cite{habernal:2016a}, {\em debate.org} \cite{durmus:2019}, and {\em reddit.com/r/changemyview} \cite{egawa:2020}. Naturally, less curation of the acquired web texts comes at the cost of more noise, which in turn calls for a cleansing of the resulting corpus.

Cleansing processes are described in several publications on argument corpora, mostly only referring to the acquired annotations though \cite{habernal:2016a,toledo:2019,gretz:2020}. In contrast, the paper at hand targets the cleansing of the corpus texts themselves. Only few works describe respective cleansing steps in detail. Among these, \newcite{alkhatib:2016a} deleted special symbols and debate-specific phrases such as ``this house'' from crawled arguments, and \newcite{habernal:2017} removed quotations of previous posts in debate posts. \newcite{wachsmuth:2017e} discarded certain types of noisy instances completely for the argument search engine args.me, but the texts in the original associated corpus \cite{ajjour:2019a} still contain much irrelevant text, as our experiments will reveal. Applying our approach has led to an improved version of that corpus. 

In this paper, we introduce a semi-supervised learning approach for corpus cleansing. In general, we follow the bootstrapping idea of successful pattern mining methods, such as {\em DIPRE}~\cite{brin:1998}, {\em Snowball}~\cite{agichtein:2000}, and {\em Espresso}~\cite{pantel:2006}. While these methods aim at semantically relevant information, we distinguish {\em pragmatically} relevant from irrelevant text within an author's argumentative discourse. We are not aware of any other approach in this direction.\,\,\,  

It is noteworthy in this regard that the cleansing task at hand differs notably from the unit segmentation of argumentative texts \cite{ajjour:2017}. While all argumentative units match the notion of relevance considered here (defined in Section~\ref{sec:approach}), also non-argumentative units may be seen as relevant, if they give facts, definitions, or other background information serving as context for the argumentative units. As such, our notion of relevance relates to the local relevance with respect to some conclusion rather than the global relevance of an argumentative statement in the discussion of an issue \cite{wachsmuth:2017b}.\,\,
\section{Approach}
\label{sec:approach}

This section presents our semi-supervised learning approach to detecting irrelevant text in web arguments as well as to clean a respective corpus on this basis. The approach aims to find as many irrelevant text units as possible at an estimated precision beyond a threshold~$\tau$ (in Section~\ref{sec:experiments}, we use $\tau = 0.95$). To this end, it learns linguistic patterns that occur often in irrelevant units and rarely in relevant units (or vice versa). Later, we consider each sentence as one unit, but other granularities would work in principle, too. Figure~\ref{process} gives an overview of the three main stages of the approach, each of which will be detailed below:%
\begin{enumerate}
\setlength{\itemsep}{0pt}
\item[(a)] 
{\em Seed Pattern Selection.} Given a corpus as input, a pool of common linguistic patterns is mined from its units, from which seed patterns indicating irrelevance and relevance are selected manually.
\item[(b)]
{\em Pattern Bootstrapping.} All units matching any seed irrelevance (relevance) pattern are retrieved, new candidate patterns are mined from the units and added to the pool. Then, only high-precision irrelevance (relevance) patterns are kept in the pool, i.e., those found nearly only in irrelevant~(relevant) units. This process is repeated until no new patterns are found or $k$ iterations have passed.
\item[(c)] 
{\em Corpus Cleansing.} The final pool of irrelevance patterns is used to automatically remove irrelevant units from the corpus.
\end{enumerate}

It is important to see that the relevance patterns are eventually {\em not} used for the actual cleansing. They serve to distinguish relevant from irrelevant units only, thereby aiding the identification high-precision irrelevance patterns. 

While we have designed our approach for web arguments in particular, notice that the outlined processed is largely generic and could easily be transferred to other cleansing tasks where relevant and irrelevant units can be distinguished. What makes our approach specific to web arguments is what we mean by argumentative relevance and irrelevance.

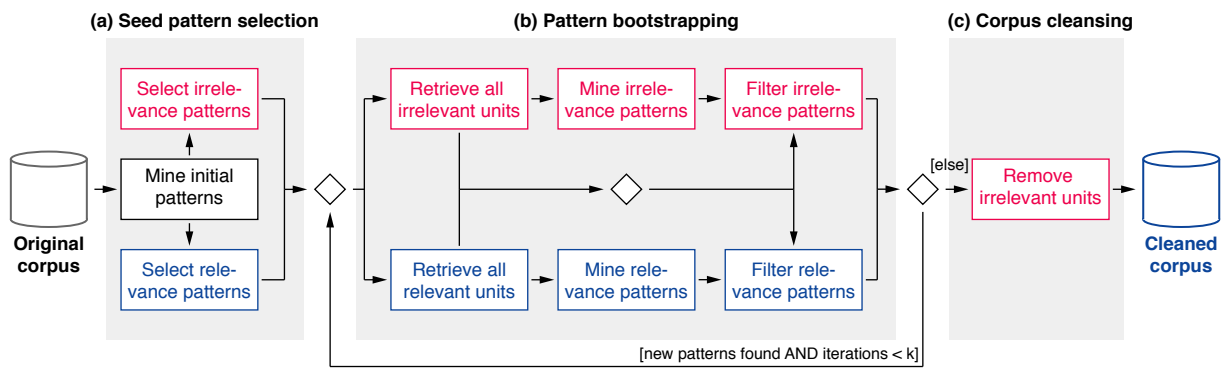
\hwfigure{process}{Conceptual process of our semi-supervised bootstrapping approach: (a)~Seed (ir)relevance patterns are selected manually from intially mined candidates. (b)~New (ir)relevance patterns are mined and filtered automatically from text units matching the existing patterns, until no new patterns are found or $k$ iterations have passed. (c)~The corpus is cleaned by removing units matching the irrelevance patterns.}

\subsection{Argumentative Relevance and Irrelevance}

We consider relevance here from the perspective of using the individual arguments in a corpus for empirical analysis of how people argue or for applications such as argument search and debating technologies. For such use cases, portal-specific debate structure emerging from sequences of arguments as well as purely rhetorical moves related to the underlying debates are not of interest. We thus define irrelevance~as~follows:\,

\paragraph{Argumentative Irrelevance.}
A unit of a web argument is said to be irrelevant, if and only if it does not represent any claim, evidence, fact, background information, or similar statement related to the issue discussed by the author of the text. Examples of irrelevant units include meta-comments on a debate, salutations, expressions of gratitude, personal insults, purely rhetorical moves, and spam.

\medskip
Any unit not matching the definition is considered to be relevant. While we could have also defined argumentative relevance instead, we decided to focus on irrelevant units, since they constitute the target concept to be detected. In other words, given that we target argument corpora, we expect irrelevant units to be the exception rather than the default. An estimation of the proportion of irrelevant units for the data processed in our experiments follows in Section~\ref{sec:data}.

\subsection{Seed Pattern Selection}

The goal of stage~(a) is to acquire a pool of linguistic patterns matching text units that can be considered either irrelevant or relevant. The set of all units matching any of these seed patterns then represents the ground-truth data that the pattern bootstrapping starts from. The selection of seed patterns is the only step that requires some level of supervision within our approach. To minimize manual effort, we propose to tackle the selection semi-automatically, i.e., we first mine the most promising candidate patterns automatically from sample data (we use a random 10\% sample of the given corpus in Section~\ref{sec:experiments}). Then, we manually classify a subset of them to be seed patterns either of irrelevance or of relevance. To do so, however, we need to first define what is considered to be a candidate pattern.

\paragraph{Candidate Patterns.}
In general, any type of linguistic pattern may be mined from corpus texts, for which respective mining methods are available. Since we expect the given notion of relevance to be largely assessable based on a unit's words only, we restrict our view to basic {\em lexical} patterns here. For simplicity, we just look for $n$-grams, but we explore four types of patterns that emerge from making two choices:
\begin{itemize}
\setlength{\itemsep}{0pt}
\item
{\em Counts vs.\ TF-IDF.} In case of counts, we simply see the $m$ most frequent $n$-grams as candidates for each $n$. In case of TF-IDF, we take those $n$-grams with the highest TF-IDF score in the sample data (each unit being one document). In our experiments, we use $m = 100$ and $n \in \{1, \ldots, 5\}$.
\item
{\em W/ stopwords vs.\ w/o stopwords.} We determine either $n$-grams based on the full unit texts (w/ stopwords) or we apply stopword removal before (w/o stopwords).
\end{itemize}
Since high TF-IDF scores usually indicate content, respective patterns are likely to be more useful for relevant than irrelevant sentences. Whether they outperform count-based patterns there is hard to predict, though. In Section~\ref{sec:experiments}, we compare the four pattern types against each other. Given all $m$ candidates of the preferred pattern type (say, {\em Counts w/o stopwords}) for each $n$, the authors of this paper then manually agree for each candidate on whether to select it as an irrelevance pattern, a relevance pattern, or neither.

\subsection{Pattern Bootstrapping}

The goal of stage~(b) is to incrementally extend the pool of irrelevance and relevance patterns using bootstrapping, i.e., by deriving new patterns from units matching the current patterns in the pool. This fully automatic process continues until no new patterns are found anymore or until a maximum number $k$ of iterations has passed, e.g., if running time is a factor (in Section~\ref{sec:experiments}, we continue until the end).

In particular, the first step is to retrieve the sets of all units matching any irrelevance patterns and of all units matching any relevance pattern from the corpus.%
\footnote{We include units that match both relevance and irrelevance patterns, since the subsequent filtering step accounts for them. Also, other performance optimizations are useful, such as storing previously found units. We leave them out here for simplicity.}
As sketched in Figure~\ref{process}, these unit sets are used for two purposes: First, new candidate irrelevance (relevance) patterns are mined from the set of irrelevant (relevant) units and added to the pattern pool. Second, only those patterns are filtered and kept in the pool that indicate an irrelevant (relevant) unit with an estimated precision $p \geq \tau$. We estimate $p$ as follows:

\paragraph{Estimated Precision.}
Let $tp$ be the number of all retrieved irrelevant (relevant) units that matches a specific irrelevance (relevance) pattern, and let $fp$ be the number of all relevant (irrelevant) units matching this pattern. Then the precision of the pattern is estimated as $p = tp \;/\; (tp + fp)$.

\medskip
For the mining step, one parameter to decide upon is the minimum frequency of a pattern to consider it a candidate. We suggest to derive this parameter's value from the seed pattern frequencies. For example, if all seed patterns have at least 20 matches in the sample, and the full corpus has 10 times the sample size, then a reasonable value may be $20 \cdot 10 = 200$. For the filtering step, it is favorable that the sizes of the two unit sets remain balanced, because imbalanced sizes decrease the comparability \nopagebreak{of the values} $tp$ and $fp$. We therefore suggest to adjust the minimum numbers based on the estimated proportion of irrelevant units. For example, if there are about 10 times as many relevant as irrelevant units, reasonable values may be 200 for irrelevance and $200 \cdot 10 = 2000$ for relevance (the numbers given here exemplarily are those we use in Sections~\ref{sec:data} and~\ref{sec:experiments}). An alternative is to test and adjust these parameters empirically.

An important characteristic of the outlined bootstrapping process is that patterns added to the pool in previous iterations may be removed later from the pool again. This is because the sets of retrieved relevant and irrelevant units change during the process, which in turn may change the precision estimations of the patterns. This can be understood as an internal revision mechanism of our approach that optimizes the precision of the final pool. We see the effect of this mechanism in our experiments in Section~\ref{sec:experiments}.%
\footnote{Depending on what sentences match the patterns, it is theoretically possible that a pattern first belongs to the relevance pool and later to the irrelevance pool (or vice versa). We did not observe notable cases in this regard, though.}

\subsection{Corpus Cleansing}

The goal of stage~(c) is to actually clean the given corpus, based on the final pool of irrelevance patterns. Relevance patterns play no role anymore in this stage; they are used only before, to be able to help identify irrelevance patterns with high precision, as described.

A simple cleansing way would be to just remove all units from the corpus that match any irrelevance patterns. Instead, however, we suggest to restrict the removal to only those irrelevant units before the first and after the last relevant unit. As long as only units are removed that are actually irrelevant, we thereby avoid to negatively affect the coherence of arguments. Moreover, as for the example of Figure~\ref{example-2}, we will see below that most irrelevant units are indeed found in the beginning and ending of texts, i.e., the suggested restriction reduces recall to some extent only. Notice that this does not mean that most units in the beginning and ending are irrelevant; in line with our discussions above, we expect the majority of texts to contain no irrelevant unit at all. The following section supports that this is true for the corpus at hand.\,

\section{Data}
\label{sec:data}

The presented approach targets argumentative language of varying quality, as often observed in web-based corpora. Below, we assess its impact on the {\em args.me corpus} \cite{ajjour:2019a}, which is to our knowledge the largest available argument corpus to this date, about 7.3~GB in file size. The corpus represents the database underlying the argument search engine args.me \cite{wachsmuth:2017e}. It contains 387,606 arguments that were mined from four debate portals using distant supervision: {\em debate.org}, {\em debatewise.org}, {\em idebate.org}, and {\em debatepedia.org}. Each argument consists of a mostly very short conclusion as well as a mostly notably longer premise, the latter containing the actual argumentative text. In total, the corpus spans around seven million sentences. We see each sentence as one unit in our approach.

Many texts in the args.me corpus include sentences that are irrelevant to the actual argument, such as the example in Figure~\ref{example-2}. Needless to say, no ground-truth information on irrelevance is given, though. For a rough estimation of the proportion of irrelevant sentences, we conducted a pilot study where the two authors of this paper independently decided about the relevance of a set of sentences, following the definition in Section~\ref{sec:approach}. In particular, we considered a corpus sample used previously by \newcite{alshomary:2020b}, which contains the top five pro and the top five con arguments each for the top~10 queries.

From the 1294 sentences in the 100 sample arguments, one of us classified 147~(11.3\%) to be irrelevant, the other one 139~(10.7\%). In terms of Cohen's $\kappa$, we had a substantial inter-annotator agreement of 0.75. In total, 175 sentences~(13.5\%) were seen as irrelevant by either of us, 111~(8.5\%) by both. Since we believe that, in doubt, a sentence should be deemed relevant, we take 8.5\% as our estimation. In the whole corpus, we thus expect around 600,000 sentences to be irrelevant. The 111 sentences come from only 39 of the 100 arguments. Assuming this number is representative, about 150k arguments in the corpus should contain irrelevant sentences. 
In the following experiments, these numbers will give us a rough idea of the recall of our approach. There, we use a random 10\% sample of all corpus arguments for the seed pattern selection, and the whole corpus for all subsequent steps.

\section{Evaluation}
\label{sec:experiments}

\begin{table*}[t!]
\small
\centering
\setlength{\tabcolsep}{7pt}
\begin{tabular*}{\linewidth}{lll@{}rcl@{}r}
\toprule		
					&				& \multicolumn{2}{c}{\bf Patterns of Relevant Sentences}			&& \multicolumn{2}{c}{\bf Patterns of Irrelevant Sentences}\\
									\cmidrule(l@{2pt}r@{2pt}){3-4}								\cmidrule(l@{2pt}r@{2pt}){6-7}																				
\bf Pattern type			& \bf n-gram		& \bf 	Most frequent pattern				& \bf Score		&& \bf Most frequent pattern			& \bf Score	 \\
\midrule
Counts				& 1-gram			& people								& 36\,287 			&& opponent						& 29\,088		\\
w/ Stopwords			& 2-gram			& the world							& 6\,210			&& my opponent					& 27\,149		\\
					& 3-gram			& the fact that							& 4\,206			&& my opponent s					& 3\,983		\\ 
					& 4-gram			& in the united states						& 977 			&& thank my opponent for				& 1251		\\
					& 5-gram			& has nothing to do with					& 494			&& i thank my opponent for			& 682		\\
\addlinespace
{\bf Counts}			& 1-gram			& people								& 36\,287 	&& opponent								& 29\,088\\
{\bf w/o Stopwords}		& 2-gram			& united states							& 3\,906			&& thank opponent 					& 1\,494		\\
					& 3-gram			& big bang theory						& 251			&& first round acceptance				& 617		\\
					& 4-gram			& life liberty pursuit happiness				& 102			&& would like thank opponent			& 359		\\
					& 5-gram			& make law respecting establishment religion	& 69	 			&& round 1 acceptance round 2		& 82			\\
\addlinespace
TF-IDF				& 1-gram			& chronicled								& 1.00 		&& ---							& ---	\\
w/ Stopwords	 		& 2-gram			& and weaponry						& 0.56			&& actually forfeiting 				& 1.00		\\
					& 3-gram			& an infinite regression					& 1.00			&& ---							& ---			\\
					& 4-gram			& abusive education and domestic			& 1.00			&& ---							& ---			\\
					& 5-gram			& acceptance of metapyhsical space that		& 1.00			&& ---							& ---			\\
\addlinespace
TF-IDF				& 1-gram			& abortions							& 1.00			&& ---							& ---	\\
 w/o Stopwords			& 2-gram			& americans like						& 1.00			&& ---							& ---			\\
					& 3-gram			& able kill others						& 1.00			&& ---							& ---			\\
					& 4-gram			& accidentally killing equivalent purposely		& 1.00			&& ---							& ---			\\
					& 5-gram			& able disprove evolution instead creationists	& 1.00			&& ---							& ---			\\			
\bottomrule
\end{tabular*}
\caption{The top $n$-gram patterns agreed upon to indicate relevant and irrelevant sentences respectively, for each evaluated pattern type, along with their score (count or TF-IDF) in the 10\% sample of the args.me corpus. We left out spam patterns, such as ``kfc ... kfc'', as they would have shadowed most other patterns. Based on the full lists (see supplementary material), we decided to use the type {\em Counts w/o Stopwords}.} 
\label{table-candidates}
\end{table*}

We now report on the step-by-step application of our approach from Section~\ref{sec:approach} to the corpus from Section~\ref{sec:data} and on the manual evaluation of the obtained results. The goal was to assess the impact of the approach on the quality of web-based argument corpora. We hypothesized that the approach is able to detect a large number of irrelevant sentences with a precision as high as its internal precision threshold~$\tau$.%
\footnote{Source code and supplementary material can be found here: \url{https://github.com/webis-de/ArgMining-20}}

\subsection{Insights into Seed Pattern Selection}

To learn what pattern type is best to detect irrelevant sentences, we compared all four candidates emerging from the two choices discussed in Section~\ref{sec:approach} (Counts vs.\ TF-IDF, w/ or w/o stopwords). For each type, we retrieved the top 100 $n$-grams, $n \in \{1, \ldots, 5\}$, covering a large variety of issues debated in the underlying arguments. Then, the two authors of this paper both judged~all~2000 resulting patterns as to whether they likely indicate always irrelevant sentences or always relevant ones. Based on the patterns that we both agreed upon, the most promising type was chosen for the seed patterns.

\begin{table*}[t!]
\small
\centering
\setlength{\tabcolsep}{2.4pt}
\begin{tabular*}{\linewidth}{lll}
\toprule																								
\bf Type 			& \bf n-grams 	& \bf 	Seed Patterns	\\								
\midrule
Relevance& 1-grams	& government (94198), states (85388), state (68123), law (59609), society (58695), money (54314), \\
                &               & death (52327), universe (49412)	\\
 \addlinespace
                & 2-grams	& big bang (9370), minimum wage (8650), human rights (8592), god exists (7959), health care (7537),\\
                &               	& years ago (7165), global warming (6399), high school (6235), opponent claims (6160),\\
                &               	& believe god (6151), human beings (6136), video games (6051), god exist (5810), existence god (5636), \\
                &               	& jesus christ (5592), supreme court (5513), new york (4890),  human life (4838), old testament (4640), \\
                &               	&years old (4632), god created (4621), god would (4495), self defense (4089), merriam webster (2015) \\
\addlinespace
		& 3-grams	& new york times (1071), world war ii (1062)\\
\addlinespace
		& 4-grams	& life liberty pursuit happiness (791), (aw respecting establishments religion (370),  \\
		&            	& make law respecting establishment (352), shall surely put death (192)		\\						
\midrule
Irrelevance& 2-grams	& first round (10113), thank opponent (10018), vote con (6048), round acceptance (4698),  \\
                	&              	& vote pro (4585), new arguments (4056), accepting debate (4040), accept debate (3432), kfc kfc (15),  \\
                	&              	&  thinking bee (3), wonyou wonyou (1), ham ham (1) \\
\addlinespace
		& 3-grams	& debate good luck (335), debate look forward (863), hi hi hi (1), dan small penis (1) \\
\addlinespace
		& 5-grams	& every one wrong every one (2)\\
				
\bottomrule
\end{tabular*}
\caption{The full lists of positive and negative and  seed patterns used for each $n$-gram type, along with the number of different sentences they match in the corpus (in parentheses), ordered by number of matches.} 
\label{table-seeds}
\end{table*}

Exemplarily, Table~\ref{table-candidates} lists the top 1- to 5-gram of each pattern type that indicate relevance or  irrelevance respectively. We left out spam patterns such as ``wonyou wonyou wonyou'' and ``kfc kfc'', though, as they would limit insights, dominating the top positions; the full lists for each pattern type are given in the supplementary material. For both {\em TF-IDF} pattern types, we find the relevance patterns to clearly serve their purpose, relating to the content of arguments. Many such patterns are found in the full lists. However, rarely any TF-IDF pattern seemed to reliably indicate irrelevance. This matches the intuition that phrases with high TF-IDF scores are specific to a document's content rather than reflecting general language. In contrast, the two {\em Counts} pattern types yielded several irrelevance patterns, as the table demonstrates. We decided for {\em Counts w/o Stopwords}, since it produced patterns that clarified many cases which {\em Counts w/ Stopwords} left ambiguous. For example, ``would like thank opponent'' reveals irrelevance knowing the source debate portals (here, debate.org), whereas respective patterns with stopwords (``would like to thank'', ``like to thank my'') leaves more doubts regarding the irrelevance of respective sentences.

Table~\ref{table-seeds} presents the full set of 38 relevance and 17 irrelevance seed patterns for the type {\em Counts w/o Stopwords}. A pattern was not included if being redundant, i.e., if it was already covered by a shorter one, e.g., ``first round acceptance'' was covered by ``first round''. We observe that no 1-gram made it into the pool of irrelevance patterns; a single word seems not enough to be sure about irrelevance. As of length~2, however, we judged several patterns to be sufficiently reliable indicators of irrelevance, the most frequent ones occurring over 10,000 times in the corpus, namely, ``first round'' and ``thank opponent''.

\subsection{Insights into Pattern Bootstrapping}

As indicated in Section~\ref{sec:approach}, we set $\tau$ to 0.95, kept all mined relevance patterns with at least 2000 matches as candidates and all mined irrelevance patterns with at least 200 matches. Given the seed patterns, we \nopagebreak{then ran} the bootstrapping process until no new pattern was found anymore, which happened in iteration~6. On a standard computer (Intel Core i7, 2.7 GHz, 16 GB RAM), the whole process took about two hours.

\begin{table*}[t!]
\small
\centering
\setlength{\tabcolsep}{9pt}
\begin{tabular*}{\linewidth}{lrrrcrrrr}
\toprule					
			& \multicolumn{3}{c}{\bf Relevance Patterns}			&& \multicolumn{4}{c}{\bf Irrelevance Patterns}						\\
			\cmidrule(l@{2pt}r@{2pt}){2-4}						\cmidrule(l@{2pt}r@{2pt}){6-9}													
\bf Iteration 	& \bf Patterns	& \bf Matches \bf 	& \bf Auto.\ Prec.	&& \bf Patterns	& \bf Matches \bf 	& \bf Auto.\ Prec.	& \bf Man.\ Prec.	\\								
\midrule
Seed			& 38			& 600\,469		& 1.00			&& 17		&	41\,619	&	0.97		& 1.00 (0.99)	\\
\addlinespace
1			& 10			& 7\,602			& 0.99			&& 74		&	5\,849	&	0.98		& 1.00 (0.96)	\\
2			& 0			& --57			& n/a				&& 19		&	3\,606	&	0.98		& 1.00 (0.94)	\\
3			& 0			& --15			& n/a				&& 4			&	956		&	0.97		& 0.96 (0.93)	\\
4			& 0			& --10			& n/a				&& 3			&	594		&	0.98		& 0.97 (0.93)	\\
5			& 0			& --6				& n/a				&& 5			&	225		&	0.98		& 0.88 (0.79)	\\
\midrule
\bf Total		& 48			& 607\,983		& 0.98			&& 122		&	52\,849	&	0.98		& 0.97 (0.92)	\\
\bottomrule
\end{tabular*}
\caption{Counts of relevance and irrelevance patterns, counts of different sentences they match, their automatically estimated mean precision, and their manually evaluated mean precision (majority agreement, full agreement in parentheses) in each iteration of our approach. The last row shows the results at the end.} 
\label{table-results}
\end{table*}

Table~\ref{table-results} shows key statistics for each iteration (and the seed pattern selection). In case of the {\em relevance patterns}, the 38 seed patterns already match more than 600k {\em different} sentences, with a mean estimated precision of 1.00, i.e., they virtually never matched any sentence retrieved for the seed irrelevance patterns. Already in iteration~2, the revision effect discussed in Section~\ref{sec:approach} starts: 57~relevant sentences were removed there, because they also matched newly mined irrelevance patterns. Still, the set of relevance patterns remained stable, and this behavior continued in subsequent iterations. For the {\em irrelevance patterns}, we observe a monotonous growth of the pattern pool in the first five iterations, with more than 10k different sentences being detected as irrelevant in iterations 1--5 in addition to the seed sentences. In total, 122 patterns were found; their mean estimated precision remained at least 0.97 in all iterations.

To analyze the behavior of our approach during the bootstrapping process, we chose a random sample of 600~irrelevant sentences for manual evaluation (found in the supplementary material): 100~matching the seed irrelevance patterns, and 100~each for the irrelevance patterns from the five iterations. Relevant patterns were disregarded, as they are not needed for corpus cleansing. We randomized the ordering of all sentences and gave them independently to three annotators with background on computational argumentation, none being an author of this paper~(one master and two PhD students; two male, one female; one each from Europe, the Middle East, and East Asia). We asked the annotators to classify each sentence as relevant or irrelevant, based on the definition from Section~\ref{sec:approach}. The annotators got some intuitive guidelines (see supplementary material) and could ask questions beforehand. 

We observed an inter-annotator agreement of 0.50 in terms of Fleiss'~$\kappa$, which seems reasonable given that relevance assessment is inherently subjective \cite{croft:2009}. Given the annotations, we computed the mean precision of our approach in detecting irrelevant sentences for each iteration, once in terms of majority agreement (irrelevance correct if two annotators say so) and once for full agreement (all three say so). The right-most column in Table~\ref{table-results} shows the results, revealing that the majority-agreement precision is perfect until the end of iteration~2. While the next two iterations remain promising, the precision decreases to 0.88 in the final iteration (0.79 under full agreement), suggesting that patterns get worse over time. An early termination may thus be favorable, but the best moment is naturally unknown in practice. 

52,849 different sentences are matched by the detected irrelevance patterns eventually, at an overall precision of 0.97. Some of them occur multiple times, resulting in 86,916 irrelevant sentences in total that come from 68,814 arguments. Under the roughly estimated irrelevance proportion from Section~\ref{sec:data}, the recall would hence be around~0.15 for irrelevant sentences and around~0.46 for arguments with irrelevance sentences. The seed step alone found 71,926 irrelevant sentences in total, i.e., a recall of roughly 0.12. If we consider the seed step as a baseline for the full approach, we see that precision decreases by 3\%~(1.00 to 0.97), but recall increases by about 20\% (0.12 to 0.15). While there is arguably room for optimization, we still conclude that the results support the impact of our approach and, by that, our hypothesis.

\subsection{Insights into Corpus Cleansing}

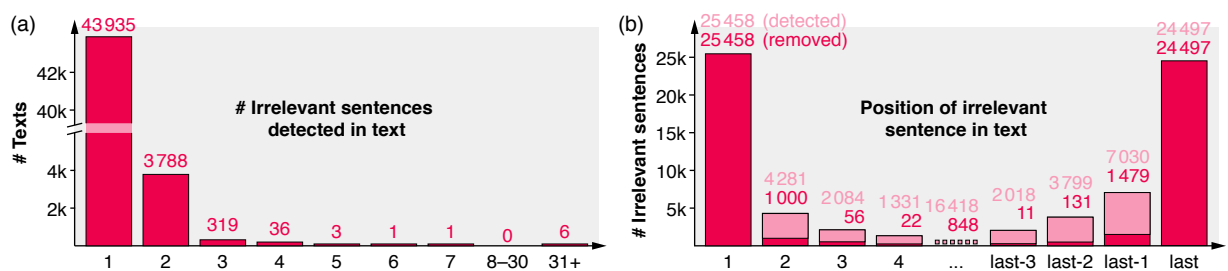
\bsfigure{irrelevant-sentences}{(a) Histograms of the number of texts in the args.me corpus with a certain a number of irrelevant sentences, as detected by our approach. (b)~Histogram of the number of detected (upper number) and removed (lower number) irrelevant sentences over the different sentence positions of a text.}

Based on the final pool of 122 irrelevance patterns, we explored the cleansing potential for the given corpus. Figure~\ref{irrelevant-sentences}(a) shows a histogram of the corpus texts with a certain number of detected irrelevant sentences. We see that most texts contain one such sentence only, in all but six cases seven or less. These six cases all have more than 30 irrelevant sentences; manual inspection revealed that they all contain spam where the same word sequence repeats itself. In Figure~\ref{irrelevant-sentences}(b), we plot the positions of irrelevant sentences in the corpus texts. As expected, most of them are found in the beginning or the end. Due to our discussed restriction of discarding only these, the final number of sentences removed from the args.me corpus sums up to 53,502 (found in 48,089 arguments). In addition to the original args.me corpus, we now also provide the cleaned corpus version at \url{https://webis.de/data.html#args-me-corpus}.

\section{Conclusion}
\label{sec:conclusion}

Web-based argument corpora play an important role in computational argumentation research and its applications. Not all text in such corpora is relevant to the arguments, though. In this paper, we have presented an approach that detects irrelevant text units in argumentative texts with low supervision. The approach iteratively bootstraps linguistic patterns of irrelevance and relevance from units matching known patterns. On the 387k arguments in the args.me corpus, the approach detected 87k irrelevant sentences at a precision of 0.97, from which at least 53k can be removed without notably reducing the arguments' coherence. These results demonstrate the potential of our approach to improve corpus quality.

Naturally, the approach has limitations. On one hand, the results revealed that, under the employed configuration, a large proportion of detected sentences came from the seed patterns. To obtain good seed patterns, manual effort is needed. On the other hand, the recall of our approach seems not so high, as far as we can estimate from the data inspected. While not all irrelevant units can be captured by the simple patterns we considered, another reason may lie in the restriction that only new candidate patterns are found which occur in sentences matching previous patterns. Particularly patterns that show up only in short units may thus be overlooked, if they are not covered by the seed patterns already. Improvements might, e.g., consider units adjacent to irrelevant units, but this may come at the cost of reduced precision. In this regard, notice that the impact our approach to some extent depends on the availability of a reliable unit boundary detector (say, a sentence splitter), which is not a trivial requirement for noisy web data.

Finally, an arising question may be how complex it is to apply the approach to other than the data processed here. Following our proposed process to obtain frequent candidate seed patterns automatically, the main manual effort boils down to finding reliable seed patterns among these candidates. In our case, this took no more than a few hours, which seems negligible given the potential impact on corpus quality. Besides, only some initial tuning of the approach parameters to the data at hand may be needed. We are thus confident that the approach can be easily adopted to clean other argument corpora (including transcribed corpora with spoken argumentative language) as well as to many other cleansing tasks where the irrelevance of text units can be defined in a measurable way.


\section*{Acknowledgments}

We thank Milad Alshomary, Wei-Fan Chen, and Jana Puschmann for their participation in the manual evaluation, and the anonymous reviewers for their helpful comments. Thank you also to Johannes Kiesel as part of the Webis Group for the technical support and the integration of the results into args.me.

\bibliographystyle{acl}
\bibliography{argmining20}

\raggedbottom

\end{document}